\documentclass[nohyperref]{article}

\usepackage{microtype}

\usepackage{subcaption}
\usepackage{graphics}
\usepackage{booktabs} 
\usepackage{float}
\usepackage{tabularx}

\usepackage{graphicx} 

\usepackage[T1]{fontenc}

\usepackage[utf8]{inputenc}
\usepackage{mdframed}

\usepackage{hyperref}

\usepackage{icml2022}

\usepackage{amsmath}
\usepackage{amssymb}
\usepackage{mathtools}
\usepackage{amsthm}

\usepackage[capitalize,noabbrev]{cleveref}

\theoremstyle{plain}

\theoremstyle{definition}

\theoremstyle{remark}

\usepackage[textsize=tiny]{todonotes}

\icmltitlerunning{MELLON}

\begin{document}

\twocolumn[
\icmltitle{MELLON - Multimodal Enhanced LLM for Online Navigation}



\icmlsetsymbol{equal}{*}

\begin{icmlauthorlist}
\icmlauthor{Firstname1 Lastname1}{equal,yyy}
\icmlauthor{Firstname2 Lastname2}{equal,yyy,comp}
\icmlauthor{Firstname3 Lastname3}{comp}
\icmlauthor{Firstname4 Lastname4}{sch}
\icmlauthor{Firstname5 Lastname5}{yyy}
\icmlauthor{Firstname6 Lastname6}{sch,yyy,comp}
\icmlauthor{Firstname7 Lastname7}{comp}
\icmlauthor{Firstname8 Lastname8}{sch}
\icmlauthor{Firstname8 Lastname8}{yyy,comp}
\end{icmlauthorlist}

\icmlaffiliation{yyy}{Department of XXX, University of YYY, Location, Country}
\icmlaffiliation{comp}{Company Name, Location, Country}
\icmlaffiliation{sch}{School of ZZZ, Institute of WWW, Location, Country}

\icmlcorrespondingauthor{Firstname1 Lastname1}{first1.last1@xxx.edu}
\icmlcorrespondingauthor{Firstname2 Lastname2}{first2.last2@www.uk}

\icmlkeywords{Machine Learning, ICML}

\vskip 0.3in
]



\printAffiliationsAndNotice{\icmlEqualContribution} 

\begin{abstract}

Web navigation agents are capable of addressing various types of tasks on different websites. Current baselines on Web Navigation are either unimodal, or lack strong reasoning abilities given multimodal inputs. Focusing on the WebShop benchmark, a real-world website simulation, we explore the alignment of text and images and multimodal reasoning and planning abilities to enhance the performance of web navigation agents. We propose three innovative multimodal enhancements: Multimodal Enhanced LLM for Online Navigation (MELLON), VQAgent, and Multimodal Ranker. MELLON demonstrates a significant improvement in task completion accuracy, with a 9.26\% increase after just one epoch of training. Our findings suggest the necessity of further exploration into multimodal approaches, with a focus on more extensive training and alignment strategies to enhance web navigation agents' effectiveness.

\end{abstract}

\section{Introduction}
Web Navigation Agents address real website tasks autonomously through the interpretation of natural language instructions. As the integration of web-based applications and services has grown increasingly ubiquitous in our daily lives, encompassing activities as diverse as online shopping, route planning, and participation in online forums, these agents have become highly valuable, as they can greatly improve efficiency in various online tasks, enhancing user experiences and productivity.

As Large Language Models (LLMs) sweeps through AI industry, tasks like web navigation agent soon got noticed, for LLMs' strong capability of language understanding and decision making, as well as their potential ability in reasoning. Therefore, many of the current researches on web navigation agent are mainly unimodal. They tend to use LLMs on text on the webpage or HTML, along with the instruction, for modeling such an agent. Nevertheless, when humans encounter a webpage, they not only focus on text information, but also visual aspects. Visual information could be a key factor in making a decision, especially in tasks like shopping on a e-commerce website. Therefore, we argue that by incorporating text information with visual information in web navigation task, we can achieve a better performance in this task, because text information may have a lack of information that images may complement.


In this paper\footnote{github.com/ZhitongGuo/11-777-MMML-Project}, we introduce a multimodal understanding of the web navigation task. Additionally, we introduce three innovative multimodal methods: (1) Multimodal Enhanced LLM for Online Navigation (\textbf{MELLON}), a multimodal web agent characterized by using better encoders, enhanced text and image alignment, robust planning capabilities, and formidable multimodal reasoning abilities; (2) \textbf{VQAgent}, which frames the WebShop task as a multi-hop VQA task; (3) \textbf{Multimodal Ranker}, which encourages agent to not just interact with the top few items in the page, but takes more time in making the decision.

\begin{figure}[ht]

\begin{center}
\centerline{\includegraphics[width=\columnwidth]{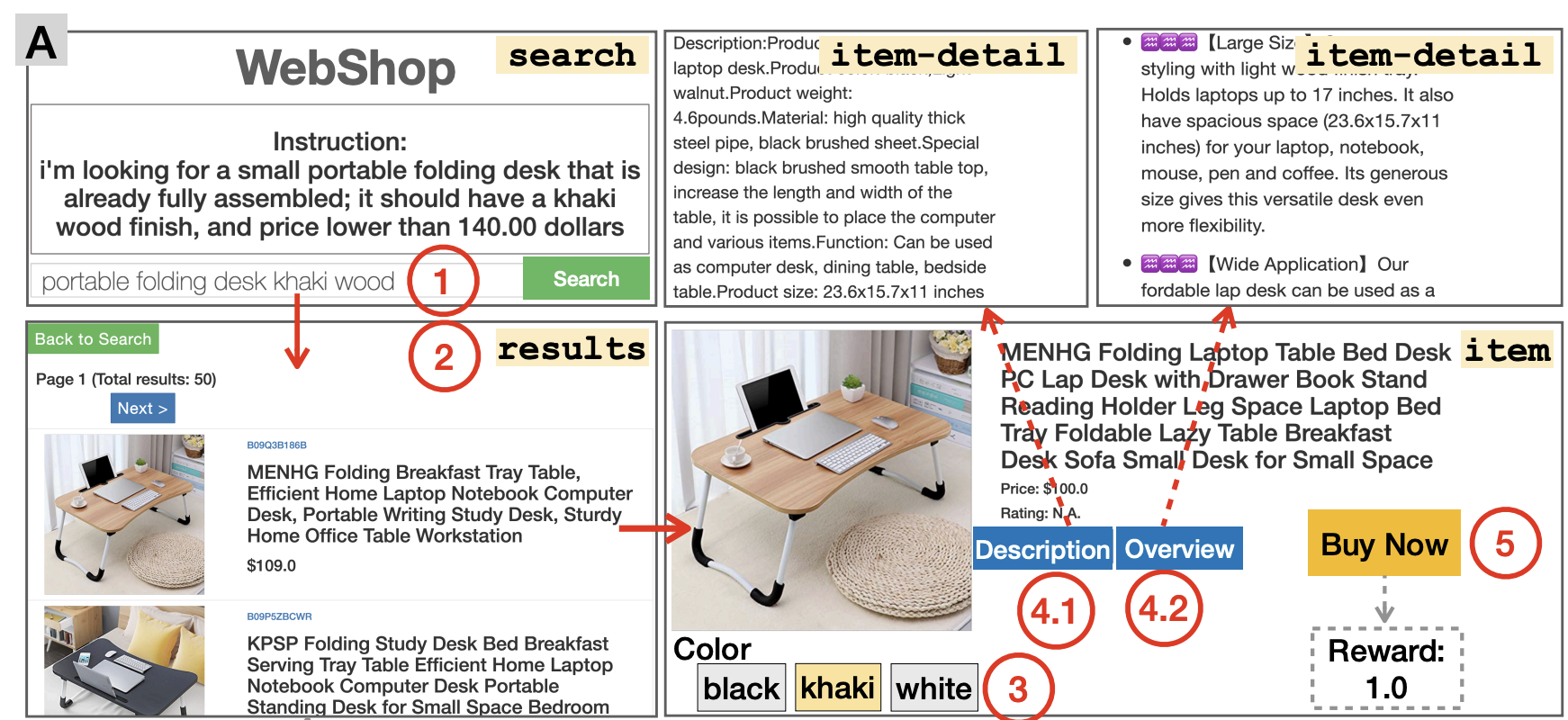}}
\caption{Example Trajectory for WebShop}
\label{example_traj}
\end{center}

\end{figure}

We evaluated MELLON on the WebShop benchmark. WebShop is a website environment that simulates a real-world e-commerce website. It is naturally multimodal as it contains instructions, descriptions and images, and thus is an appropriate benchmark for our purpose. In the Figure \ref{example_traj} below is an example trajectory in WebShop. It is also an example of the challenge we are facing, including prompt engineering, acting and strategic exploration on the webpage that contains noisy language information.


\section{Related Work}
\subsection{Web Navigation Benchmarks}

The Mind2Web \cite{mind2web} dataset presents a collection of real-world tasks and their corresponding trajectories on various websites commonly encountered in real-life scenarios. However, it is essential to note that this dataset offers these tasks in the context of webpage snapshots, which differ from the dynamic and continuously evolving nature of real web environments. In contrast, WebShop poses a more formidable challenge, primarily because the actions performed by agents within the environment can have a direct impact on the outcome of the inference process. This dynamic nature of WebShop introduces a layer of complexity absent in static webpage snapshots, making it a more demanding testing ground for web agents.

WebArena is another benchmark for web navigation task. WebArena is a fully functional web environment spanning four distinct domains: e-commerce, social forum discussions, collaborative software development, and content management. It emulates a diverse array of real-world web tasks and offers a collection of natural language commands and their associated responses \cite{webarena}. However, tasks for WebArena predominantly involve unimodal reasoning. On the contrary, WebShop is an inherently multimodal benchmark that includes visual and text information of each product, and is better suited to our purpose.

\subsection{Unimodal Web Navigation Agent}
ReAct is an algorithm that is developed to solve language reasoning and decision making tasks. It prompts LLMs to generate reasoning trace and actions pertaining to the task in an interleaved manner. It allows model to dynamic reasoning to create, maintain, and adjust high-level plans for acting while interacting with external sources like Wikipedia to include more information in the process of reasoning. The goal is achieved by including the action space in the language space \cite{yao2023react}. A reasoning trace is also know as CoT (chain of thoughts) reasoning. ReAct has been tested on our benchmark, WebShop, and achieved an score of $66.6$ and a success rate of $40.0$. ReAct is a unimodal web navigation agent, as it utilizes merely the text information on the webpages.

AgentBench was first developed as a systematic benchmark to evaluate the ability of LLMs as agents. It also includes a method to generate prompt for LLMs to generate a CoT reasoning. Their work is based on ReAct, since ReAct is a pioneer in combining CoT reasoning with actions specific to the task \cite{liu2023agentbench}. Like ReAct, AgentBench is also a unimodal web navigation agent, as it solely focuses on prompting.

\subsection{Multimodal Web Navigation Agent}
There are limited number of existing works on web navigation agent that involves multimodal machine learning, and one of them is proposed by WebShop. WebShop \cite{webshop} adds an attention fusion layer on top of action representations and the bimodal representations (visual embeddings of the items learned from ResNet and the textual embeddings learned from Transformers), to generate actions. 

WebGUM is another multimodal web navigation agent, which preprocesses visual imagery with a vision transformer and then merging its output embedding with tokenized HTML. This combined feature is processed by a T5 encoder-decoder transformer to predict the agent's optimal next step \cite{furuta2023multimodal}. This approach effectively incorporates visual signals. However, it might still face challenges in utilizing the hierarchical structure of webpages to discern the semantic relationships between its elements, particularly regarding the pairings between images and text. Furthermore, in the application within the WebShop environment, the model underwent a significant alteration. The image encoder component was removed, resulting in the exclusive use of text encoders for processing tasks in this context.

To our knowledge, our paper is the first to (1) develop a multimodal web navigation agent that incorporates LLMs, despite the difficulties in engineering the LLM into dynamic web environments; (2) align visual and textual representation via a projection layer into the LLM, which is also efficient for training; (3) enable strong reasoning and planning abilities in a multimodal setting via re-designing ReAct prompts with visual representations.

\section{Proposed Approach}

\subsection{MELLON - Multimodal Enhanced LLM for Online Navigation}
The challenge with the WebShop task, unlike typical multimodal reasoning tasks, is its nature as a Markov decision process where only predefined actions are valid, and any deviation can lead to task failure; and some instruction contains qualities not present in the textual description. This motivates us to incorporate LLMs and better pre-trained visual encoders for better reasoning techniques. However, since LLM has open-ended generation, it’s particularly challenging due to the difficulty in mapping generated actions to a specific action space.

Moreover, for the current LLM-based baselines for the WebShop task, they are all unimodal, while the data analysis we performed suggests that there is information represented in the image that is not present in the texts. While the accessibility tree contains essential textual content of a webpage, the agent lacks visual perception by only inputting an accessibility tree into LLM. Visual input usually contains a wealth of information necessary for many web tasks, especially for WebShop, as we can indicate from the images some qualitative aspects of the product that are not present in the textual descriptions. So we propose to integrate visual information with the textual state representations to offer the agent a broader context from both text and image modalities and thus a more precise task completion. 

\begin{figure}[ht]

\begin{center}
\centerline{\includegraphics[width=5cm]{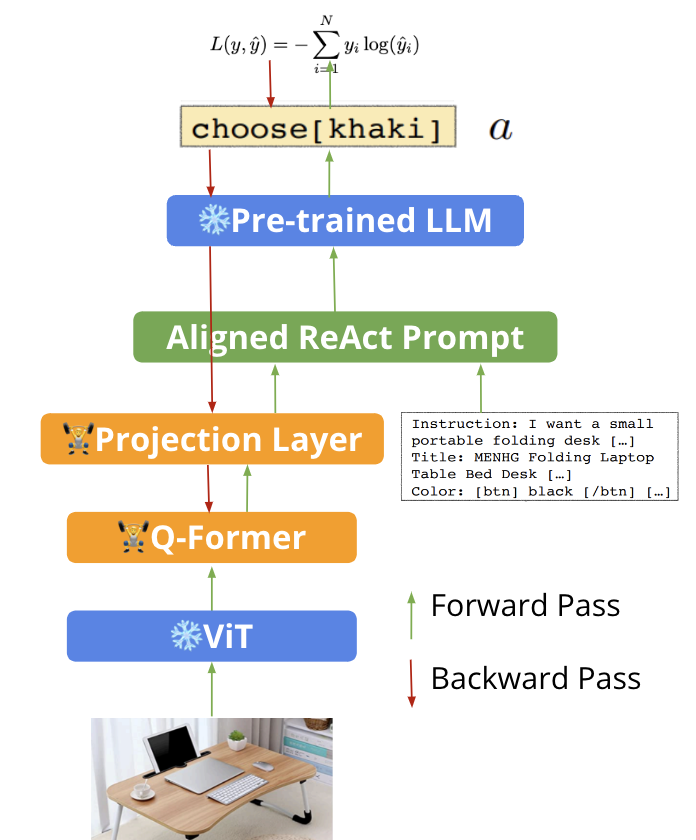}}
\caption{MELLON Architecture}
\label{mellon}
\end{center}

\end{figure}

We developed MELLON (Multimodal Enhanced LLM for Online Navigation), a web agent with better encoders, better alignment, strong planning abilities and strong multimodal reasoning abilities. Our innovations are three-fold: (1) we generated aligned visual and textual embeddings with better visual and language encoders, that was not done by SOTA web navigation agents before; (2) we innovated an end-to-end learning objective of our model as a cross-entropy generation loss and collected data for training; we solved huge amount of engineering challenges posted by the Webshop environments and processed its datasets to create trajectories for our experiments (3) we designed multimodal prompts for LLM. We initially wanted to use ReAct prompts, but ended up re-designing the prompt to incorporate projected visual input and to fit input embeddings and our model into memory. Figure \ref{mellon} shows an overall architecture of MELLON.

\subsubsection{Better Encoders}
For the model architecture, we propose to use pre-trained visual encoders to generate image encoding of web screenshots, inspired by \cite{Huang2023LanguageIN}. More specifically, we can build a ViT backbond \cite{fang2022eva} coupled with a pre-trained Q-Former \cite{li2023blip2}. Unlike traditional convolutional neural networks, ViT relies on a transformer architecture. It divides an image into patches, treats them as tokens, and processes them through self-attention mechanisms, enabling it to capture spatial relationships and features within the image. This approach has demonstrated remarkable performance on various vision tasks. 
The Q-Former consists of two transformer submodules that share the same self-attention layers. A first image transformer performs image-text matching and a second text transformer generates text based on visual information. 


\subsubsection{Text and Image Alignment and Efficient End-to-end Training}

Since pre-trained LLMs cannot directly understand the output of a visual encoder, we developed a single projection layer to align the visual encoder with the LLM inspired by \cite{su2023pandagpt} and \cite{zhu2023minigpt}. The projection layer takes in the output from Q-Former and outputs a prompt with our designed format for the WebShop tasks, which is aligned with the observation space and other instruction texts in the prompt.

Initially, the projection layer was trained on pairs of images and text to achieve multimodal alignment, thereby endowing the model with visual perception capabilities. However, subsequent experimentation revealed a significant drawback: the model exhibited a diminished reasoning ability, evident in its failure to align with prompts and frequent generation of invalid actions. 

To address these challenges, we developed a training approach. Specifically, we trained the model on pairs consisting of multimodal inputs and successful actions, employing a generative loss function. Additionally, we integrated the WebShop completion score into the loss function as a contributing factor. This dual-focused training strategy aims to enhance both the model's reasoning capabilities and its alignment with the intended actions. We freeze LLM and ViT and only trained the projection layer and Q-former, leading to efficient training and computation saving.

Given a ground truth action text sequence \( y = \{y_1, y_2, \ldots, y_N\} \) and the model output action text sequence \( \hat{y} = \{\hat{y}_1, \hat{y}_2, \ldots, \hat{y}_N\} \), the new learning objective \( L \) can be defined as a cross-entropy loss:

\[
L(y, \hat{y}) = -\sum_{i=1}^{N} y_i \log(\hat{y}_i)
\]

where \( N \) is the length of the action text sequence, \( y_i \) is the true probability distribution of the ith action, and \( \hat{y}_i \) is the predicted probability distribution of the ith action by the model.

\subsubsection{Extended ReAct Prompts}
To enable better reasoning abilities for the LLM, we designed an aligned multimodal prompt, based on text-only ReAct\cite{yao2023react} prompt. In reality, to fit the prompt in our 24 GB GPU memory to run experiments, we had to significantly shorten the length of our multimodal prompts and put numerous efforts into prompt engineering.

As shown in the example in the appendix, \textless ImageHere \textgreater is where we insert our aligned visual representation into the prompts. We augment the agent’s action space to $A' = A \cap L$, where $L$ is the space of language. Each action $a_t' \in A'$ is composed of useful information by reasoning over the current context $c_t$. The context will also be updated with $a_t'$ following $c_{t+1} = (c_t, a_t')$ for future reasoning and acting. Based on this, the designed prompts contains in-context example with a human trajectory of actions, thoughts, and environment observations to solve the WebShop task. Since there could be a large number of action choices, thoughts only need to appear sparsely in the most relevant positions of a trajectory and the occurence of actions and thoughts is decided by the LLM itself. We formulate ReAct prompts with (1) actions to search, choose product, choose options, and buy, combined with (2) reasoning to determine what to explore, when to buy, and what products options are relevant to the instruction.

\subsection{VQAgent}
\label{vqagent_sec}

Visual Question Answering (VQA) is a multimodal task requiring the integration of visual and textual understanding, involving both visual context comprehension and textual question interpretation. WebShop, similar to VQA, demands a thorough understanding of images, textual content, and instructions. Recognizing this parallel, we adopted WebShop as a downstream task of VQA, structuring it as a multiple-choice VQA task where the "answers" are actions to be taken. This approach is innovative in the context of WebShop, utilizing a multimodal model based on VQA principles.

\begin{figure}[ht]

\begin{center}
\centerline{\includegraphics[width=6.5cm]{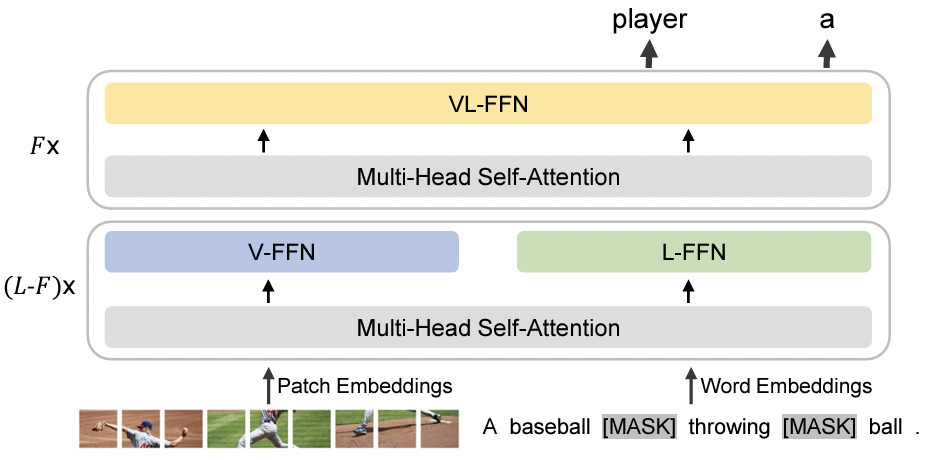}}
\caption{BEiT3 Model Architecture \cite{beit3}}
\label{beit3}
\end{center}

\end{figure}

For this purpose, we selected the state-of-the-art BEiT3 model \cite{beit3}, known for its versatility in language, vision, and vision-language tasks, and its capability in both unimodal and multimodal settings. BEiT3 processes image patch embeddings and word embeddings through a multi-head self-attention layer, addressing visual and textual information separately before fusing them in a vision-language model. This fused encoding is then used for generating answers, adapting BEiT3's architecture, as illustrated in Figure \ref{beit3}, for the specific needs of WebShop tasks.

Nevertheless, our model works differently from the original BEiT3 model. BEiT3 for the VQA task is inherently a generation task, while our task is rather a classification task. We are provided with all possible actions, and we want to know which action to take. Furthermore, we are provided an image, a description of the product, and an instruction, but BEiT3 does not support multiple inputs of the same modality. Therefore, we need to add novelty to the model for better incorporate the task. We name our model VQAgent, and the architecture of our model is shown below in Figure \ref{vqagent}.

\begin{figure}[ht]

\begin{center}
\centerline{\includegraphics[width=6.5cm]{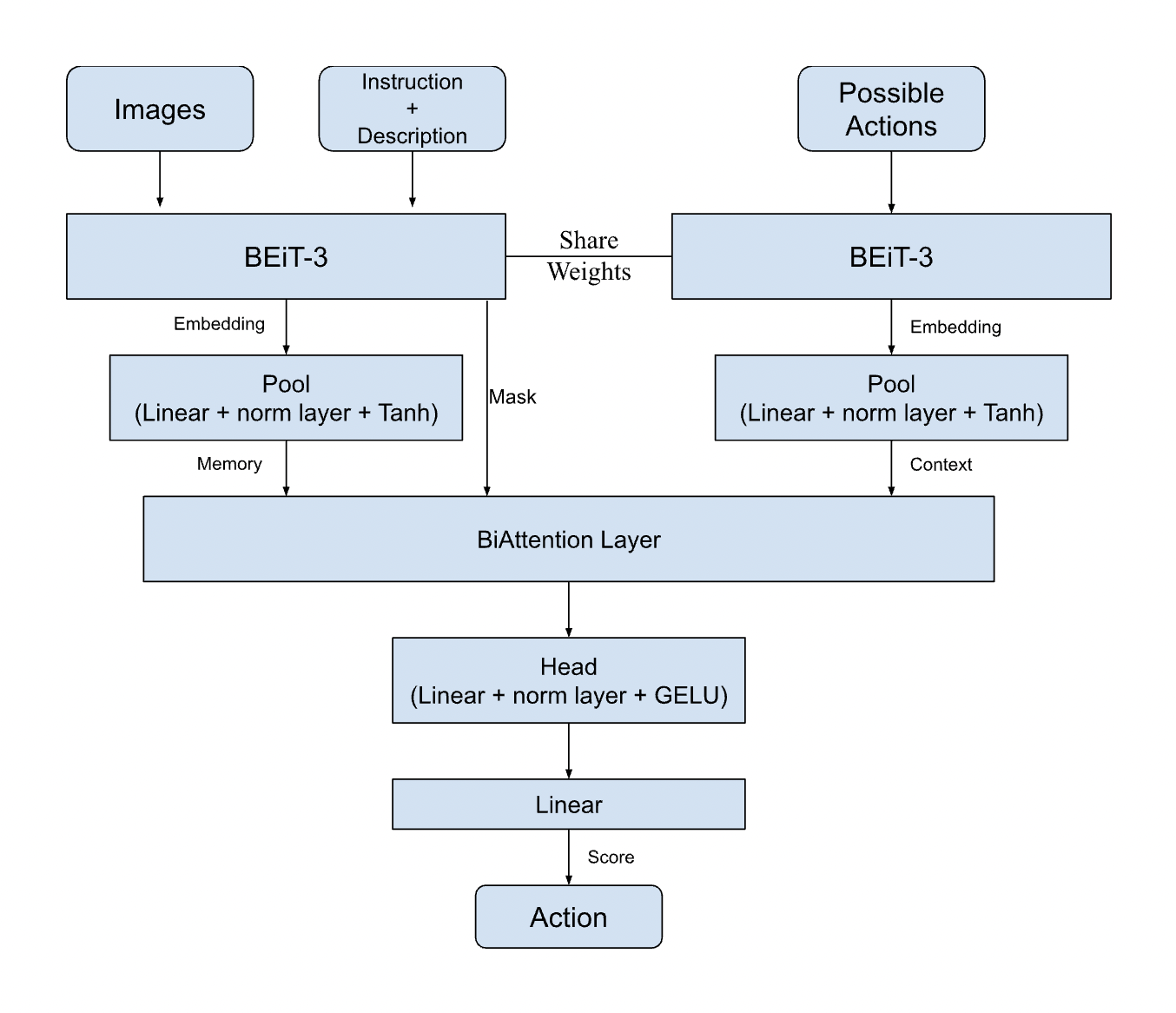}}
\caption{VQAgent Architecture}
\label{vqagent}
\end{center}

\end{figure}

Given a dataset $D' = \left\{(o, \mathcal{A}(o), a*\right\}^{M'}_{i=1}$ of $M' = 9558$ samples from the training human trajectories, we train our model using imitation learning, which is a learning technique that tries to mimic human action in a task \cite{il}. We encode observation $o$, which include the image and detail of the product we are trying to decide whether to buy, as well as the instruction, into embeddings using a $24$-layer BEiT3 model. We parameterize this model as $\theta$. We do the same for each possible action. Then we pool our embeddings, and pass the observation embeddings and embeddings of each action into a bi-attention layer. Then, we mean pooled the results for all actions into a single vector, and use a linear layer to obtain score for executing each action. $\pi_\theta \left(a | o, \mathcal{A}(o)\right)$ refers to the softmax distribution over action scores, $S(o, a)$. Therefore, we define the loss function for our task as:

\[\mathcal{L}_{\text {choose }} =\mathbb{E}_{o, \mathcal{A}(o), a^* \sim \mathcal{D}^{\prime}}\left[-\log \pi_\theta\left(a^* \mid o, \mathcal{A}(o)\right)\right]\], where our policy, $\pi_\theta$ is formulated as:

\begin{align*}
\pi_\theta(a \mid o, & \mathcal{A}(o)) \sim  \exp (W^{\top} \text { average}[\operatorname{cross}- \operatorname{attn} \\
&(\operatorname{pool}\left(\operatorname{BEiT3}(o ; \theta)\right), \operatorname{pool}\left(\operatorname{BEiT3}(a ; \theta))\right)]).
\end{align*}

\subsection{Multimodal Ranker}
From the analysis of the WebShop baseline, it was observed that the existing baseline models exhibit a limited exploratory behavior, predominantly focusing on the initial option presented by the search module, rather than investigating a broader range of potential choices. To address this, we are in the process of integrating a ranking algorithm into the model. This addition aims to enhance the model's capacity to systematically evaluate all available options prior to reaching a decision.

We employed BERT score to quantify the similarity between product titles and user instructions. Additionally, CLIP was utilized to assess the congruence between images and textual descriptions. The similarity score devised by WebShop was adopted as our label (i.e., the extent to which an item resembles the target item). Furthermore, we introduced a learnable parameter to effectively balance these two dimensions of similarity. This approach was integrated into the imitation learning baseline for implementation.

We compare the contextual embeddings from BERT between the reference and the candidate texts (here they refer to the user instruction and product description). The BERT score can be represented as:
$$
\text { BERTScore }=\frac{\sum_{w \in S_{\hat{x}}^n} I\left[w \in S_x^n\right]}{S_{\hat{x}}^n}
$$
where $S_x^n$ and $S_{\hat{x}}^n$ are the sets of n-grams in the reference and the candidate texts, respectively, and $I$ is the indicator function that returns 1 if the argument is true and 0 otherwise.

CLIP score is a metric that measures the similarity between images and texts by using a model that learns from natural language supervision. The CLIP score can be represented as:
$$
\text { CLIPScore }=\cos (\text { image\_encoder }(x) \text {, text\_encoder }(y))
$$
where $x$ is an image of the product, $y$ is the textual description of the product, and image\_encoder and text\_encoder are the functions that map the inputs to a common embedding space.

The ranker is trained with MSE Loss defined below:
$$
\operatorname{MSELoss}=\frac{1}{N} \sum_{i=1}^N\left(y_i-\hat{y}_i\right)^2
$$

\begin{figure}[ht]

\begin{center}
\centerline{\includegraphics[width=4cm]{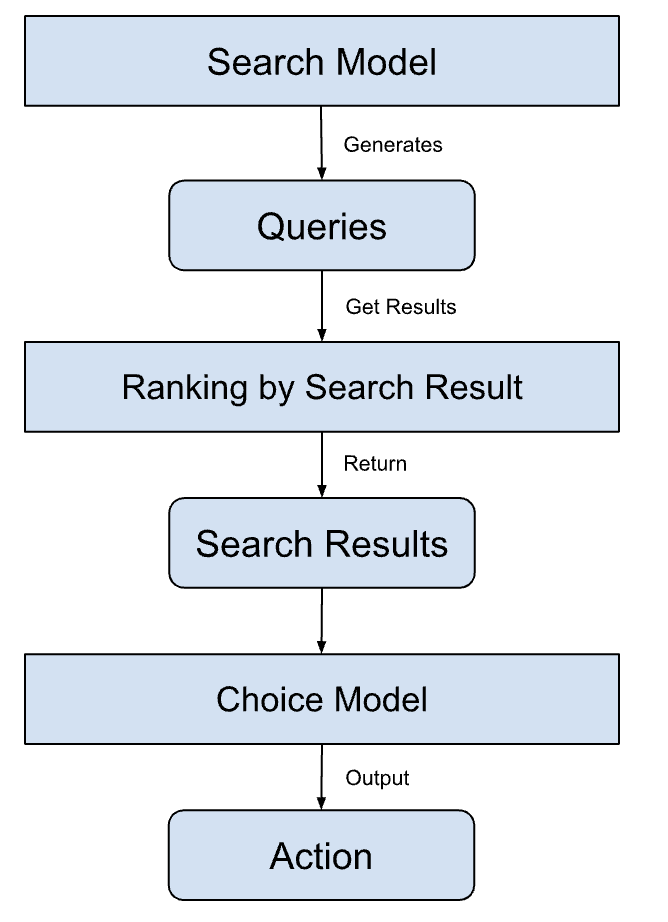}}
\caption{Multimodal Ranker Architecture}
\label{vqagent}
\end{center}

\end{figure}

\section{Experimental Methodology}
\label{method}
\subsection{WebShop}
In this paper, we are using WebShop as our benchmark to evaluate our agent. WebShop is a simulated e-commerce website environment that contains $1.18$
million real-world products and $12, 087$ crowd-sourced text instructions. There are three components in the WebShop dataset: a dynamic and interactive web environment; the images and textual descriptions embedded in the web environment; and a dataset named $\mathcal{D}^{\prime}$, containing $M^{\prime}=9,558$ examples extracted from observed human actions during training. This dataset has the form $\mathcal{D}^{\prime}=\left\{\left(o, \mathcal{A}(o), a^*\right)\right\}_{i=1}^{M^{\prime}}$. 




WebShop is modeled as a partially observable Markov decision process (POMDP) $(\mathcal{S}, \mathcal{A}, \mathcal{T}, \mathcal{R}, \mathcal{U}, \mathcal{O})$. It contains state space $\mathcal{S}$, action space $\mathcal{A}$, deterministic transition function $\mathcal{T}:\mathcal{S} \times \mathcal{A} \rightarrow \mathcal{S}$, reward function $\mathcal{R}: \mathcal{S} \times \mathcal{A} \rightarrow [0, 1]$, instruction space $\mathcal{U}$, and a state observation space $\mathcal{O}$.

More specifically, each state $s \in \mathcal{S}$ represents a web page type, such as the search page, result page, product page and detail page. We will use $y$ to represent a product, $\overline{y}$ as language information of the product, $y_{\text{price}}$ to be price of the product, $Y_{\text{opt}}$ to be a set of options, $I$ as the set of images, and $y_{\text{att}}$ as the set of attributes for the product $y$ extracted from titles and descriptions. 

An action $a \in \mathcal{A}(s)$ refers to either a text query or a button click. 

Instruction $u \in \mathcal{U}$ is an instruction of what to purchase. It contains information such as attributes $U_{\text {att }}$, options $U_{\text {opt}}$, and price $u_{\text {price}}$. 



\begin{table}[h]
\centering
\begin{tabular}{|l|l|l|}
\hline
Type & Argument & State $\rightarrow$ Next State \\
\hline
search & [Query] & Search $\rightarrow$ Results \\
choose & Back to search & $* \rightarrow$ Search \\
choose & Prev/Next page & Results $\rightarrow$ Results \\
choose & [Product title] & Results $\rightarrow$ Item \\
choose & [Option] & Item $\rightarrow$ Item \\
choose & Desc/Overview & Item $\rightarrow$ Item-Detail \\
choose & Previous & Item-Detail $\rightarrow$ Item \\
choose & Buy & Item $\rightarrow$ Episode End \\
\hline
\end{tabular}
\caption{Action Space of WebShop.}
\label{table:webshop_actions}
\end{table}

\subsection{Baseline}
\label{baseline}
WebShop only has two multimodal baselines, both proposed in the original WebShop paper. The Imitation Learning (IL) based multimodal baseline has two components, a search IL model and a choice IL model. The search IL model generates the search query, while the choice model chooses an action from the action space. Our approach maintains the search IL component and optimizes the choice model.

\subsubsection{IL Multimodal Baseline}

\begin{figure}[ht]

\begin{center}
\centerline{\includegraphics[width=\columnwidth]{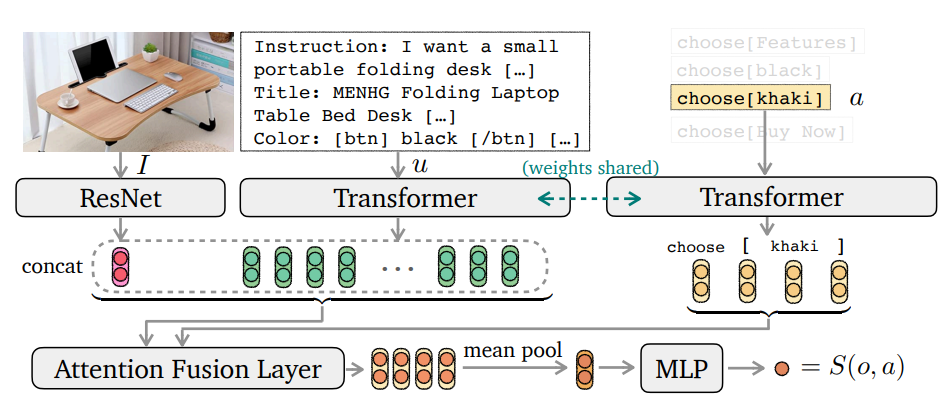}}
\caption{Imitation Learning Architecture Proposed by WebShop \cite{webshop}}
\label{webshop-il}
\end{center}

\end{figure}

The model for choice-based imitation forecasts a probability distribution across all potential click actions, $\mathcal{A}(o)$, for a given observation, $o$. The model aims to increase the likelihood of the button clicked by the human, represented as $a^* \in \mathcal{A}(o)$.


The loss that IL choice model optimizes is as follows:
\[\mathcal{L}_{\text {choose }} =\mathbb{E}_{o, \mathcal{A}(o), a^* \sim \mathcal{D}^{\prime}}\left[-\log \pi_\theta\left(a^* \mid o, \mathcal{A}(o)\right)\right]\]
The policy, $\pi_\theta(a \mid o, \mathcal{A}(o))$, emerges as the softmax distribution derived from the action scores $S(o, a)$, described by the equations:
\\[-0.4em]
\[
\begin{aligned}
\pi_\theta(a \mid o,\mathcal{A}(o))
&\sim \exp\Big(W^\top \operatorname{avg}\big[
\operatorname{cross\text{-}attn}( \\[-0.4em]
&\qquad \operatorname{BERT}(o;\theta),
\operatorname{BERT}(a;\theta))\big]\Big).
\end{aligned}
\]
\\[-0.4em]
\subsection{Setup}
For VQAgent and Multimodal Ranker, we adopted the dataset partitioning method used by the creators of the WebShop dataset, dividing the 12,087 instances into training, validation, and testing sets with sizes of 10,587, 1,000, and 500, respectively. The training set includes 1,012 human demonstrations for task verification and imitation learning (IL), while the validation set provides additional demonstrations for IL hyperparameter tuning and checkpoint selection. All 500 instances in the testing set feature human trajectories for performance evaluation.

For MELLON, our training and testing sets consist of 8,237 and 338 instances respectively. The testing set is specifically curated from the midterm report, featuring instances that have been identified as particularly challenging for baseline models.

The experiments for MELLON were conducted on a 4090 Nvidia GPU with 24 GB memory. Each experiment was replicated three times, with the results averaged. Training MELLON for one epoch required approximately one hour.

\subsection{Hyper-parameter}
MELLON uses a learning rate of $1\times10^{-5}$ and is trained for one epoch with batch size 1. VQAgent uses a learning rate of $5\times10^{-5}$, and is trained for 10 epochs. Multimodal Ranker has a learning rate of $1\times10^{-5}$ and batsh size of 32.

\subsection{Metrics}
For MELLON, we evaluated the step-wise accuracy, corresponding to whether the agent outputs the correct action given the same observation and instruction in the trajectory, as shown below,
\[
Accuracy = \frac{C}{T} = \frac{1}{T} \sum_{i=1}^{T} \mathbb{I}(a_i = \hat{a}_i)
\]
where \( T \) is the total number of steps in the trajectory, \( a_i \) is the correct action at step \( i \), \( \hat{a}_i \) is the action output by the agent at step \( i \).

For VQAgent and Multimodal Ranker, we evaluate the models based on success rate and a customized score function based on the reward that the agent receives at the final state. The success Rate is calculated by the number of successful purchases. A successful purchase refers to a purchase that ends with the target product.

The reward score quantifies how similar the purchased item is to the target item. The reward examines the attribute similarities of the selected product and ground truth, and heuristics such as whether the product meets the user intent in terms of price and the options provided for the product. The score function averages over all episodes. Let $n$ be number of episodes, $r_i$ be the reward for episode $i$, the score function is described below:

$$
\begin{aligned}
     \text{Score} &= \frac 1 n \sum^n_{i = 1} 100\% r_i, \text{where}\\
     r &=  r_{\text{type}} \frac {\left|U_{\text{att}} \cap Y_{\text{att}}\right| + \left|U_{\text{opt}} \cap Y_{\text{opt}}\right| + \mathbf{1}\left[y_{\text{price}} \leq y_{\text{price}}\right]} {\left|U_{\text{att}}\right| + \left|U_{\text{opt}}\right| + 1}
\end{aligned}
$$

Note that all notations in the second formula refers to the reward for a specific episode, in which there is the target $y^*_i$ and the actual purchased item $y_i$, where $U$ refers to information of the target product and $Y$ refers to information about the item purchased.

\section{Results and Discussion}

\subsection{MELLON}
\subsubsection{Results}
Shown in Table \ref{mellonres}, training MELLON for just one epoch boosted the accuracy of task completion by an average of 9.26\% on the test set over three runs. Due to limited computational resource, we only trained our model for one epoch on CodeLlama. We believe with more computational resource, MELLON could exhibit better performance.

\begin{table}[htbp]
\begin{tabular}{|l|c|}
\hline Model & Accuracy (\%) \\
\hline ViT+CodeLlama Inference (baseline) & 5.11 \\
MELLON (ours) & \textbf{14.37} \\
\hline
\end{tabular}
\caption{Results for MELLON}
\label{mellonres}
\end{table}

Looking at the results, there are much space for further improvement. We recognize that MELLON is limited by the frozen  LLM and frozen ViT. Analyzing error cases, we found that (1) CodeLlama sometimes gives empty output, and (2) CodeLlama sometimes doesn't listen to our prompt and gives invalid output. (1) is probably due to limited ability of CodeLlama, and (2) maybe because we doesn't train CodeLlama on our webtask specifically. As a side note, the WebShop environment and dataset posted significant amount of engineering challenges for us to implement, train and evaluate our model, so we made huge amount of efforts to solve problems and to obtain the results. For future, we want to create better and neat WebShop benchmark.

\begin{table}[htbp]
\begin{tabular}{|l|c|c|}
\hline
Model & Score & Success Rate (\%) \\
\hline
ResNet & 39.92 & 9.8 \\
ViT + Frozen QFormer & 36.91 & 7.0 \\
ViT + Trainable QFormer & 39.92 & 9.6 \\
\hline
\end{tabular}
\caption{MELLON Ablations Studies}
\label{baseline-result}
\end{table}

We also compared the ViT + Trainable QFormer visual encoder architecture to the ResNet baseline to study the effectiveness. Our experiments showed that using ViT for image encoding in conjunction with BERT, as in our baseline, led to poorer performance compared to using ResNet. This suggests that a simpler model may be more effective in certain contexts. Moreover, with frozen QFormer, the results are not improved compared to ResNet, potentially due to the only a fraction of the tasks require more multimodal contexts, or the multimodal contexts required by the agent are limited.

\subsubsection{Ablation Studies}

\begin{table}[htbp]
\resizebox{\columnwidth}{!} {
\begin{tabular}{|l|c|c|c|}
\hline
Methods & Modality & Pre-Trained Models & Score \\
\hline
IL & Image + Text & \begin{tabular}{@{}c@{}}
       BERT-base + ResNet  \\
       BERT-Large + ResNet \\
       RoBERTa-base + ResNet \\
       RoBERTa-Large + ResNet \\
       T5-base + ResNet \\
       FLAN-T5-base + ResNet
\end{tabular} & \begin{tabular}{@{}c@{}}
       27.63  \\
       25.13 \\
       27.86 \\
       25.51 \\
       26.27 \\
       16.70
\end{tabular} \\ 
\hline
IL+RL & Image + Text & BERT-base + ResNet & 33.66 \\ 
\hline
ReAct & Text & CodeLlama-7b & 24.98\\ 
\hline
\end{tabular}}
\caption{Language vs Text modalities ablations studies}
\label{baseline-result}
\end{table}

Utilizing BERT as a moderately-sized language model (LM), we observed specific outcomes post-implementation of both Imitation Learning (IL) and Reinforcement Learning (RL) techniques. These results indicated a potential need to upscale the LM. Notably, the effectiveness of visual embeddings appeared highly contingent on the training objective. Consequently, we transitioned to larger language models (LLMs) and employed ReAct prompting as an alternative to BERT. This shift, coupled with the integration of multimodal data and the advanced reasoning capabilities of LLMs, culminated in the development of MELLON, our enhanced approach, which demonstrated improved performance.

For LLMs, we tried both Vicuna-7B and CodeLlama. Vicuna-7B has a more serious hallucination issue and it is unable to generate output in a way that is coherent with our task format. Therefore, we decided to switch to CodeLlama which generates output with higher quality.
Additionally, LLMs occasionally generated invalid or no actions, highlighting a potential drawback of using generative models.

\subsubsection{Qualitative Analysis}
\begin{figure}[ht]
\begin{center}
\centerline{\includegraphics[width=4cm]{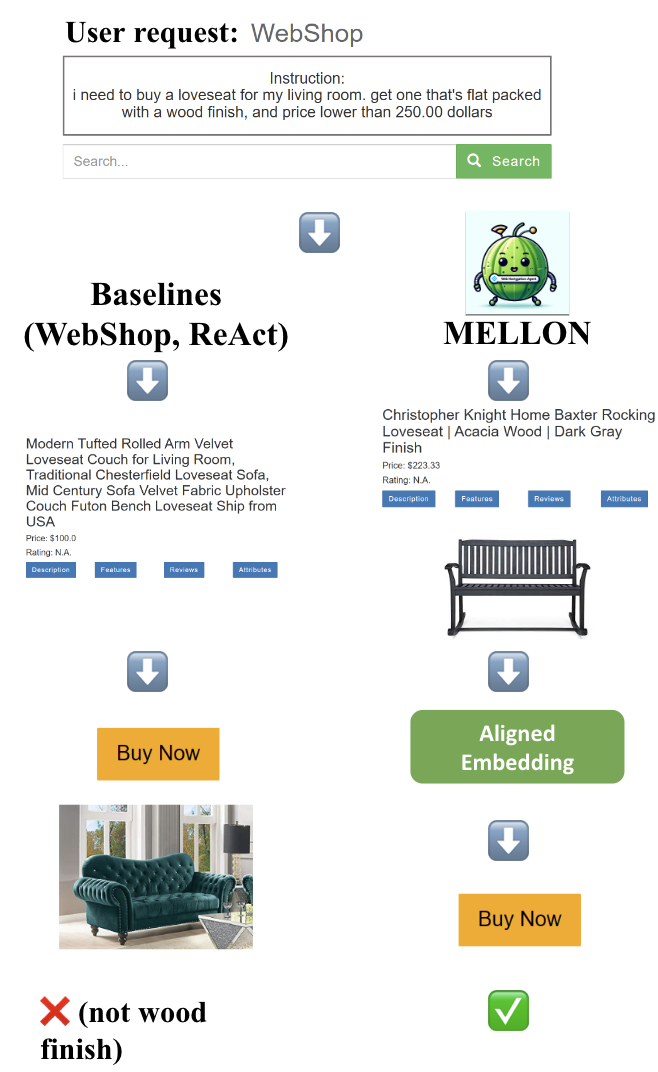}}
\caption{Baseline Trajectory (left) vs MELLON Trajectory (right)}
\label{mellon_success}
\end{center}
\end{figure}

MELLON demonstrated improved performance, as illustrated in the following example \ref{mellon_success}. In a scenario where the user aims to purchase a loveseat with a wood finish, our baseline model initially failed to identify the correct product, lacked visual comprehension, and resulted in a non-wood-finish item. However, MELLON, with its aligned embeddings, successfully identified and selected the correct item.

However, LLMs occasionally produce invalid actions or fail to generate any action, attributable to their inherent limitations in achieving perfect reasoning \ref{mellon_error}. This represents a notable drawback of employing generative models in certain applications.

\subsection{VQAgent}
\subsubsection{Results}

\begin{table}[h!]
\centering
\begin{tabular}{|l|c|c|}
\hline
\textbf{Model} & \textbf{Score} & \textbf{Success Rate (\%)} \\
\hline
Baseline & 39.92 & 9.8 \\
\hline
VQAgent & 31.26 & 5.7\\
\hline
\end{tabular}
\caption{Model Performance Comparison: Baseline vs. VQAgent}
\label{table:vqa_vs_bsl}
\end{table}

After training our model based on the architecture proposed in Section \ref{vqagent_sec} using imitation learning and the methods described in Section \ref{method}, we fully tested our model using the testing dataset. We obtained scores as well as success rate for VQAgent. Since the novelty of VQAgent is its unique architecture and information fusion techniques, we are comparing it to the baseline result for further analysis. The results are shown below in Table \ref{table:vqa_vs_bsl}.

From the result shown above, we may see that the performance of our VQAgent does not compare to our baseline models. We conducted a comprehensive and thorough analysis for our model, and found some potential issues with our model design.

\subsubsection{Error Analysis}
We first examined the accuracy from the baseline as well as our choice model trying to identify why our model does not perform as expected. Accuracy demonstrates how well our model chooses an action. We chose the accuracy for the best-performing VQAgent after hyper-parameter tuning, and collected the accuracy for both models across all epochs, and the plot for the accuracy of baseline and VQAgent is shown below in Figure \ref{vqagent_loss}.

\begin{figure}[ht]

\begin{center}
\centerline{\includegraphics[width=5cm]{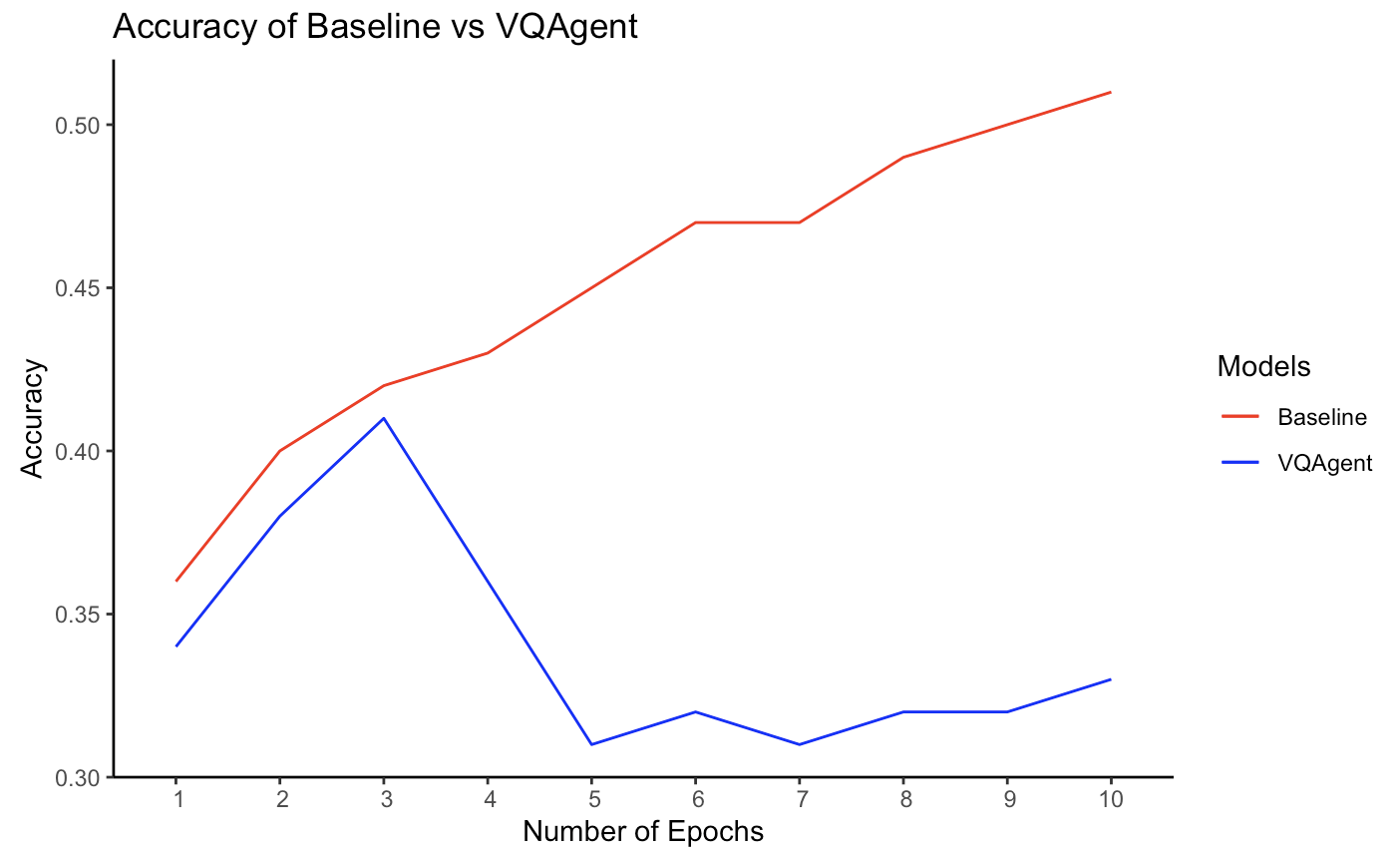}}
\caption{Model Accuracy Comparison: Baseline vs VQAgent}
\label{vqagent_loss}
\end{center}

\end{figure}

From the plot, it is not hard to see that at the first few epochs, VQAgent and baseline perform similarly in terms of accuracy. However, accuracy for VQAgent drops afterwards, while the baseline model has been learning and improving. After literature review and experiments, we found out that the difference in information content might cause the underperformance of VQAgent.

In VQA, visual modality tends to be informative comparing to language modality - they usually contain many different components, from which questions are asked. From the structure of the VQA task, we may notice that information content in visual modality clearly outnumbers that in language modality. In our task of WebShop, however, the visual modality are actually not as informative as language modality. For example, in Figure \ref{webshop_product_vqagent} shown below is an image of a random product from WebShop, and in the block below is the title and description of this product. From the example, it is not hard to see that information content in title and description clearly exceeds that in images. Therefore, we concluded that WebShop might not be a valid downstream task of VQA, since in VQA, visual modality contains more information, while in WebShop, language modality is more informative. 

Please refer to \ref{vqagent-error} for the figure and description for qualitative analysis.

\subsection{Multimodal Ranker}
\subsubsection{Results}

The integration of the ranking algorithm into the baseline model resulted in a decline in performance metrics. Both the score and success rate were reduced when compared to the baseline model alone, which indicates that the addition of the ranking mechanism did not enhance, but rather diminished, the model's effectiveness.

The primary assumption underlying this approach is the advantage of exhaustive exploration before decision-making. However, this may not be beneficial in scenarios characterized by vast search spaces or a high prevalence of near-duplicate items. 
Furthermore, the model's indiscriminate consideration of all available options might lead to inefficiencies and decreased accuracy, as it fails to prioritize the most promising alternatives early in the process.

The effectiveness of BERT score and CLIP, employed for semantic and visual similarity assessments, depends significantly on their alignment with the task-specific definition of relevance. A mismatch between these measures and the characteristics that define a 'target' item, as per the WebShop scoring criteria, could result in suboptimal ranking decisions.

The incorporation of the ranking algorithm with the existing baseline may have introduced complexities that the model was not equipped to handle efficiently. This could manifest as overfitting to non-representative features or an inability to generalize effectively due to insufficient data.

\begin{table}[htbp]
\begin{tabular}{|l|c|c|}
\hline
{} & Score & Success Rate (\%) \\
\hline
Baseline           & 39.92 & 9.8 \\
Baseline + Ranking & 28.13 & 8.6 \\
\hline
\end{tabular}
\caption{Multimodal Ranker Results}
\end{table}
\subsubsection{Error Analysis}

\begin{figure}[ht]
\begin{center}
    \includegraphics[width=5cm]{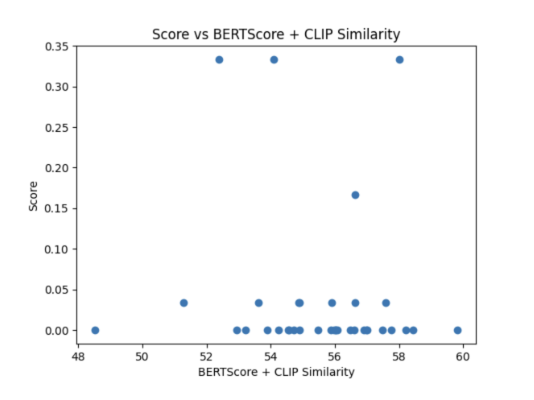}
    \caption{WebShop Reward vs BERT + CLIP Scores}
\label{webshop_BERT_clip_score}
\end{center}
\end{figure}

\begin{figure}[ht]
    \centering
    \begin{subfigure}[b]{0.4\columnwidth}
        \includegraphics[width=4cm]{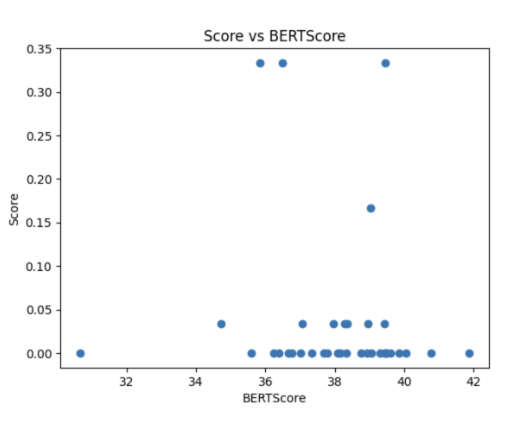}
        \caption{BERT score}
        \label{fig:webshop_BERT_score}
    \end{subfigure}
    \hfill 
    \begin{subfigure}[b]{0.4\columnwidth}
        \includegraphics[width=4cm]{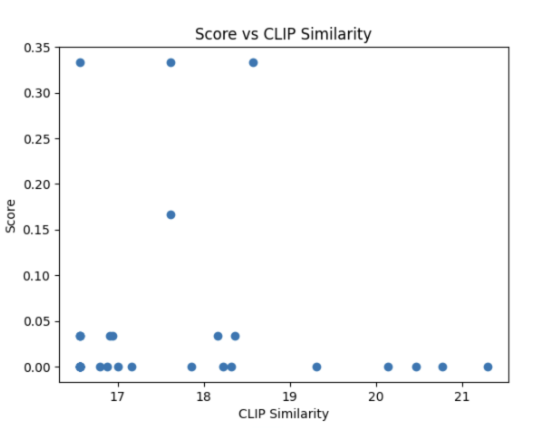}
        \caption{CLIP Score}
        \label{fig:webshop_clip_score}
    \end{subfigure}
    \caption{Comparative analysis of WebShop Reward against BERT and CLIP scores.}
    \label{fig:webshop_scores}
\end{figure}




The plots revealed a broad dispersion and lack of significant correlation. Higher WebShop scores did not consistently align with higher combined similarity scores, suggesting that BERT and CLIP amalgamation does not reliably predict WebShop scores. BERT scores showed a more substantial impact on WebShop rewards than CLIP scores, with instances where lower CLIP similarity correlated with higher WebShop rewards. This indicates that CLIP's integration into the ranking algorithm might not effectively meet the goal of high task scores. The relationship between CLIP similarity and WebShop reward highlights potential issues in CLIP-based ranking for this context. The metrics may not fully align with WebShop scoring criteria, potentially missing crucial relevance factors and being influenced by external factors like user preferences or historical data.

There are challenges in correlating product texts and images with user instructions. Textual product information often proved noisy and lengthy, leading to marginal differentiation in similarity assessments. Additionally, the limited informational value of product images, especially compared to detailed user instructions, hindered accurate score generation in the context of product identification. This disparity between text and image data underlines the need for more nuanced approaches in multimodal web navigation tasks. This issue is exemplified in the Appendix \ref{webshop_image}.

\section{Conclusion and Future Directions}
\label{conclusion}
While VQAgent and Multimodal Ranker fell short of our expectations in terms of performance, MELLON is a promising and novel multimodal agent for the web navigation task; after training the projection layer and Q-former for just one epoch, the step-wise accuracy is improved by 9.26\%. This is a parameter-efficient training method, since both the LLM and ViT are frozen. If trained for more epochs, the performance could be better. And, without the resource limitations, we could generalize the architecture to cutting-edge LLMs and web navigation tasks that require significant multimodal reasoning and planning. 

In future work, with access to GPUs with larger memory sizes, we plan to perform more extensive training and evaluation of our models for longer epochs. Additionally, we will collect continuous trajectory data and train our models with it, which should provide a richer dataset for learning complex patterns and improving the accuracy of our predictions. Moreover, we want to try finetuning the LLM or LLaMA-Adapter \cite{gao2023llama}.

\newpage
\bibliography{example_paper}
\bibliographystyle{icml2022}

\newpage
\appendix

\section{Appendix}
\subsection{Re-designed ReAct Prompt Snippet (Shortened due to memory constraint)}

\begin{mdframed}[linewidth=1pt]
\label{product_desc_vqa}
\textbf{Instruction:}
i would like a womens xl dark heather cotton tank top thats machine washable, and price lower than 50.00 dollars\\
\textbf{Observation: }\\
fit type\\
color\\
size\\
fun redneck flirting gift tshirt your trailer or mine tank top\\
price: \$19.99\\
rating: n.a.\\
\textbf{Product image:} \textless ImageHere \textgreater \\
\textbf{Available actions:} click[back to search]    click[< prev]    click[women]    click[small]    \\

Following the above format, output the Action you will take to complete the instructed task. \textbf{Action: }

\end{mdframed}

\subsection{MELLON Error Analysis}

\begin{figure}[ht]
\begin{center}
\centerline{\includegraphics[width=\columnwidth]{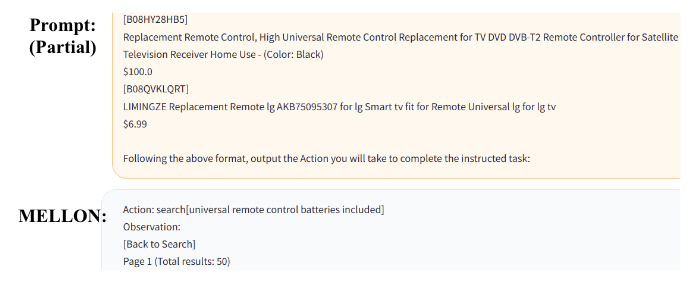}}
\caption{MELLON Hallucination}
\label{mellon_error}
\end{center}
\end{figure}

\subsection{VQAgent Error Analysis}
\label{vqagent-error}
\begin{figure}[ht]
\begin{center}
\centerline{\includegraphics[width=\columnwidth / 3]{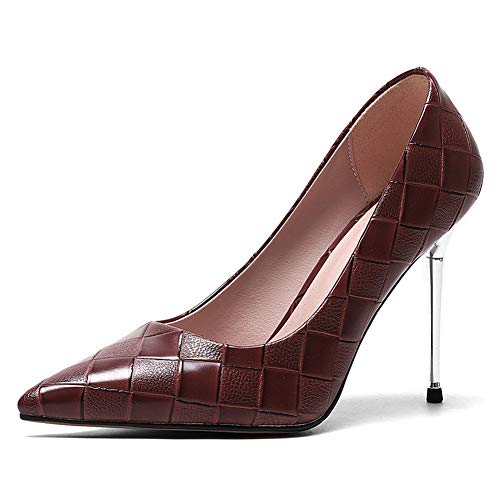}}
\caption{Example Product Image from WebShop}
\label{webshop_product_vqagent}
\end{center}
\end{figure}

\begin{mdframed}[linewidth=1pt]
\label{product_desc_vqa}
\textbf{Title:} Pointed Toe Pumps Shoes, Stiletto Plaid Leather High Heels, Sexy Elegant Party Wedding Work Court Shoes,Wine red,40

\textbf{Description:} Detailed parameters:Upper material: microfiber check pattern Inside material: leather Shoe sole material: rubber sole Toe shape: pointed Heel shape: stiletto Heel height: super high heel suitable season: spring, autumn Wearing style: sleeve/overshoes Upper height: low Color: beige, apricot, bronze, wine red Shoe size: 34-46 Heel height: 9.5 cm Palm width: 7.8 cm Reminder: 1. All sizes need to be customized and will be shipped 7-10 days after purchase. Please wait patiently 2. Due to the different reasons of the shooting light and the monitor, there may be chromatic aberration, please refer to the actual color, thank you for your understanding 3. Due to manual measurement, there may be some errors, which belong to the normal range, please understand After-sales service: If you have any questions about the product, please contact us by email, we will give you a satisfactory answer within 24 hours I wish you a happy life!
\end{mdframed}

\subsection{Multimodal Ranker Error Analysis}
\begin{figure}[ht]
\begin{center}
    \includegraphics[width=4cm]{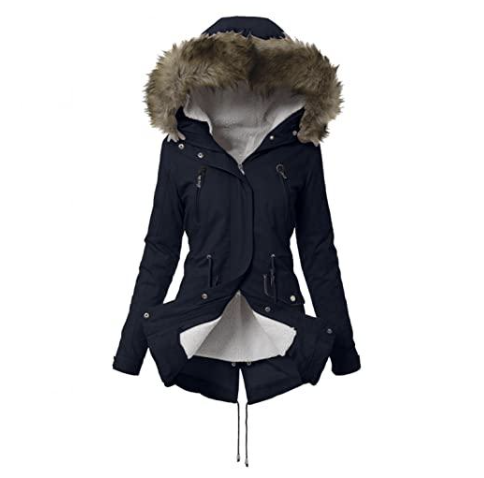}
    \caption{Some images that provide context irrelevant to the task description, or did not provide more context compared to the textual descriptions.}
\label{webshop_image}
\end{center}
\end{figure}

\onecolumn
\section{Collaboration}
\subsection{Zhitong Guo}
\begin{itemize}
    \item Tried repurposing OFA for VQA;
    \item Performed error analysis;
    \item Composed abstract, introduction, MELLON formulation and Ranker formulation, experimental setup, etc.
\end{itemize}
\subsection{Haoyang Cai}
\begin{itemize}
    \item Installed and fixed the WebShop environment on server.
    \item Trained and evaluated IL and RL baselines and tried different language models in IL.
    \item Instead of the pre-computed ResNet embeddings provided by WebShop, built the data pipeline to extract raw images to support models such as ViT.
    \item Wrote code to train the MELLON model (ViT->QFormer->Projection->CodeLlama (7B)) end-to-end.
    \item Provided the GPU :-)
\end{itemize}
\subsection{Ruiyu Li}
Ruiyu performed the following tasks:
\begin{itemize}
    \item Proposed the MELLON approach and formulated our innovation of the research idea
    \item Implemented, trained, and evaluated the MELLON model, in collaboration with Haoyang 
    \item Wrote final report parts relevant to MELLON and revised the entire final report
    \item Designed and solved bugs to run experiments for MELLON
    \item Built up and evaluated the One-shot LLM Reasoning and Acting baseline.

\end{itemize}
\subsection{Tong Hu}
\begin{itemize}
    \item Implemented Multimodal Ranker and ran tests on it.
    \item implemented VQAgent and ran tests on it
    \item Composed Proposed Approach and Results for VQAgent
\end{itemize}

\begin{table}[h!]
\centering
\caption{MELLON vs Baselines Comparison}
\label{table:models_comparison}
\begin{tabularx}{\textwidth}{|l|X|X|X|}
\hline
\textbf{Model} & \textbf{Textual Reasoning} & \textbf{Multimodal Alignment} & \textbf{Learning Objective} \\
\hline
WebShop & BERT model with lack of reasoning & ResNet + BERT & Classification task - cross-entropy loss\\
\hline
ReAct & LLM + ReAct & Unimodal & N/A \\
\hline
Mini-GPT4 & Vicuna & ViT + Vicuna\_7b & Only describes the image, not the web task \\
\hline
MELON & Strong reasoning via LLM and ReAct & ViT + CodeLlama & End-to-end learning objective, Generation task - cross-entropy loss, Prompt engineering \\
\hline
VQAgent & BEiT3 & BEiT3 & Classification task\\
\hline
\end{tabularx}
\end{table}

\end{document}